\documentclass[sigconf]{acmart}

\copyrightyear{2026}
\acmYear{2026}
\setcopyright{cc}
\setcctype{by}
\acmConference[DAC '26]{63rd ACM/IEEE Design Automation Conference}{July 26--29, 2026}{Long Beach, CA, USA}
\acmBooktitle{63rd ACM/IEEE Design Automation Conference (DAC '26), July 26--29, 2026, Long Beach, CA, USA}
\acmDOI{10.1145/3770743.3807726}
\acmISBN{979-8-4007-2254-7/2026/07}

\usepackage{booktabs}
\usepackage{graphicx}
\usepackage{amsmath}
\usepackage{xcolor}

\begin{document}

\title{Late Breaking Results: Audio-Language-Action Models with Direct Speech Conditioning for Robotic Manipulation}

\author{%
Enfu Nan\textsuperscript{1},
Pu Zhao\textsuperscript{1}$^*$, 
Yixiao Chen\textsuperscript{1,2},
Lin Zhao\textsuperscript{1}, \\
Juyi Lin\textsuperscript{1},
Chen Wang\textsuperscript{2},
Weiwei Chen\textsuperscript{2},
Yanzhi Wang\textsuperscript{1,2}
}

\affiliation{%
  \institution{%
    \begin{tabular}{cccc}
      \textsuperscript{1}Northeastern University &
      \textsuperscript{2}EmbodyX Inc. 
    \end{tabular}
  }
  \city{}
  \country{}
}

\renewcommand{\shortauthors}{Nan et al.}
\begin{abstract}
Vision-language-action (VLA) models enable robots to follow 
instructions in typed text, limiting broader deployment in natural speech.
The traditional solution by prepending a speech recognition before VLA to convert speech into texts would incur additional latency and propagate errors. 
To address this, we present \textbf{Speech-$\pi_{0.5}$}, which maps raw speech directly to motor commands without explicit textual transcriptions. Specifically, built on $\pi_{0.5}$, we use  a cross-modal projection to convert voice frames directly into audio tokens with significant response latency reduction. To further improve the accuracy, we adopt \emph{split LoRA} adaptation with dedicated audio and task adapters, under a two-stage training. 
Speech-$\pi_{0.5}$ achieves $3\times$ lower response latency than  ASR  pipeline while maintaining competitive task success on LIBERO, and generalizes  robustly to unseen accents where a single adapter  degrades significantly.
\end{abstract}

\keywords{vision-language-action models, speech-conditioned robot control, multimodal learning,  foundation models, audio-language-action}

\maketitle

\section{Introduction}

Vision-language-action (VLA) models~\cite{black2024pi0,pi05,brohan2023rt2,kim2024openvla}
have made rapid progress in enabling robots to follow natural language
instructions, but they universally require \emph{typed text} as input.
This may  limit their wide deployment in practical natural speech environments \cite{zhao-etal-2024-pruning,shen2024search,zhan2024exploring,zhan2024fast,shen2025sparse,shen2024lazydit}.
The traditional solution is to prepend an automatic speech recognition (ASR)  such as Whisper~\cite{radford2023whisper}, which converts speech into texts first, and then feeds texts to   VLA.
However, it incurs large response latency to trigger robot actions due to   verbose textual transcription, with potential   accuracy loss from transcription errors.

Although recent works~\cite{vlas,tang2024salmonn}  show that 
speech tokens can be integrated into  LLMs via projection, 
integrating speech into high-performance VLAs such as $\pi_{0.5}$~\cite{pi05} remains largely 
unexplored, as their   vision-language backbones present a modality gap that cannot be bridged by projection alone.
To address this, we propose \textbf{Speech-$\pi_{0.5}$}, %
with three key contributions:
(i)~cross-modal projection %
that compresses speech %
into compact VLM tokens;
(ii)~\emph{split LoRA} %
that improve audio quality and accent robustness over a standard single adapter; 
and
(iii)~two-stage training to prevent catastrophic forgetting. 
Our model eliminates the ASR decoder, achieving $3\times$ lower response latency with competitive
task success on LIBERO~\cite{liu2023libero}.

\section{Method}

\begin{figure}[t]
  \centering
  \includegraphics[width=0.95\linewidth, trim=0 0 0 0, clip]{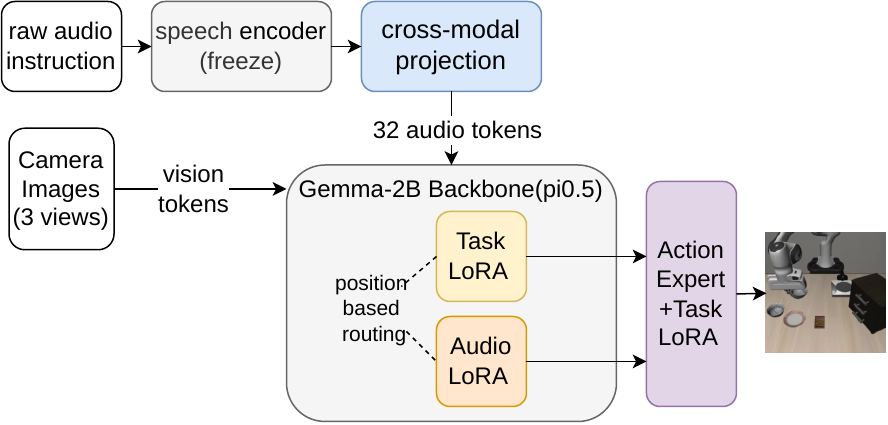}
  \vspace{-10pt}
  \caption{Speech-$\pi_{0.5}$: a frozen speech encoder's output is projected
 into 32 VLM  tokens. Split LoRA adapts  backbone with
  separate audio and task adapters routed by token position.
  }
   \vspace*{-10pt}
  \Description{Architecture diagram showing audio processed by a frozen speech encoder, projected and pooled into the VLM prefix with split LoRA backbone adaptation.}
  \label{fig:arch}
\end{figure}

We propose  \textbf{Speech-$\pi_{0.5}$}, which accepts voice instructions and outputs robotic actions for robot tasks. Built on $\pi_{0.5}$, our method has three key techniques (Figure~\ref{fig:arch}), including (i) cross-modal projection, (ii) split LoRA adaption, and (iii) two stage training.

\subsection{Cross-Modal Projection}
We design a cross-modal projection to convert speech hidden states into VLM tokens. Specifically, 
given a spoken instruction, a frozen pretrained speech encoder produces
frame-level hidden states $\mathbf{H} \in \mathbb{R}^{N \times d_s}$
($N{=}1500$ frames). Then a two-layer MLP with GeLU activation projects these into
the VLM token space ($\mathbb{R}^{d_v}$). A \emph{learned cross-attention
pooling module} with $K{=}32$ learnable query vectors then compresses the
projected sequence:
{\small\[
\mathbf{Z}_\text{audio} = \text{CrossAttn}(\mathbf{Q},\,\text{MLP}(\mathbf{H}))
\in \mathbb{R}^{32 \times d_v}.
\]}%
where $\mathbf{Q} \in \mathbb{R}^{32 \times d_v}$ is a set of 32 learnable query vectors optimized during end-to-end training.
By compressing 1500 speech frames into 32 tokens via attention pooling, we bypass the autoregressive ASR decoder, reducing ASR-stage latency by 94–97\%.

\subsection{Split LoRA  Adaptation.}
To equip the model with audio capabilities, we adopt LoRA~\cite{hu2022lora} to finetune the VLM backbone. However, we observe that a single LoRA is easy to overfit to training voices, with sharply degrading accuracy on unseen accents (\S\ref{sec:ablation}). To address this,  
we therefore propose to use \textbf{split LoRA}:
each of the 18 Gemma-2B backbone layers receives \emph{two} rank-16
adapters---a \textbf{task LoRA} ($\mathbf{B}_t\mathbf{A}_t$) active at
text/action positions and an \textbf{audio LoRA} ($\mathbf{B}_a\mathbf{A}_a$)
active at audio positions. For hidden state $\mathbf{h}_i$ at position $i$,
the LoRA output is:
{\small
\begin{equation}\label{eq:splitlora}
\Delta\mathbf{h}_i =
  \begin{cases}
    \mathbf{B}_a \mathbf{A}_a \, \mathbf{h}_i & i \in \mathcal{P}_\text{audio} \\
    \mathbf{B}_t \mathbf{A}_t \, \mathbf{h}_i & i \in \mathcal{P}_\text{text/action}
  \end{cases}
\end{equation}}%
Position-based routing keeps the two pathways separate. %
This separation ensures that speech adaptation does not interfere with task learning, preserving task performance while enabling robust generalization across diverse audio commands.

\subsection{Two-Stage Training}

\textbf{Stage~1: Speech--Language Alignment.} 
To effectively supervise ASR, we use the  outputs of the original PaliGemma ($\pi_{0.5}$ backbone)  under the ground-truth text inputs, to distill our speech-$\pi_{0.5}$ with our projection  and audio LoRA adapter. By aligning the  outputs of the original model under texts and our model under audio through MSE loss distillation, our modules are enforced to extract and align linguistic content from audio tokens. 
Training uses robot-domain TTS covering 100k instructions across ${\sim}30$ voices, mixed with task-specific vocabulary.
The action expert remains frozen.

\noindent\textbf{Stage~2: Task Training.}
We optimize the LoRA adapters (both the task-specific and audio-specific), the action expert LoRA, and the audio projector, using the flow matching action loss on robotic demonstration data, while keeping the speech encoder and base model weights frozen. Crucially, the task and audio LoRA parameters are updated at different learning rates, with the audio LoRA assigned a smaller learning rate. This asymmetric update scheme prevents task-oriented gradients from overwriting the learned audio representations, thereby preserving speech understanding capability throughout manipulation training. The joint optimization of the projection layers and LoRA modules is essential: it enables the model to acquire task-level manipulation competence while retaining the generalization ability of the speech recognition pathway.

\section{Experiments}\label{sec:ablation}

We evaluate on \textbf{LIBERO-Spatial} and \textbf{LIBERO-Object} (10 tasks
each, 50 episodes per task, $n{=}500$ per condition)~\cite{liu2023libero}.
We use Whisper-large-v3~\cite{radford2023whisper} as the frozen speech encoder.
We use Piper TTS (22 VCTK speakers) to train audios. For held-out evaluation, we use 10 Azure Neural TTS voices with accents unseen during training.

\noindent\textbf{High Accuracy.} Our method achieves higher action accuracy than single LoRA, as split LoRA adaptation mitigates interventions from audio tokens  for robotic tasks.
As shown in Table~\ref{tab:main},  our method reaches 89.2\% on LIBERO-Spatial and 86.0\% on LIBERO-Object, higher than single LoRA with non-marginal improvements.

\noindent\textbf{Accents Robustness.} With the proposed LoRA Adaptation and two stage training, our  Speech-$\pi_{0.5}$  achieves higher accents robustness. As shown in Table~\ref{tab:main}, when our finetuned Speech-$\pi_{0.5}$ is evaluated on held out accents  unseen during training, the accuracy of  LIBERO-Spatial and  LIBERO-Object  only drops slightly. By contrast,  on LIBERO-Object, the single LoRA baseline without our split loRA adaptation suffers from significant accuracy drop from 69\% to  38\% due to overfitting. Differently, by isolating audio representations in a dedicated
adapter, our method learns accent-invariant speech features rather than overfitting to training voices.

\noindent\textbf{Response Latency Reduction.} 
Our method can significantly reduce the response latency for the first action, since our model directly addresses voice tokens without  voice decoding into plain texts. 
As shown in Table~\ref{tab:latency}, traditional ASR pipeline needs to decode voice into texts with massive computations, resulting in significant latency before the first action. By contrast, our cross-modal projection  converts speech features into tokens which are sent directly to following VLMs, skipping the steps of decoding into texts and then further encoding into tokens.  With our method, the first action fires in 109 ms, achieving a 3$\times$ reduction in response latency.

\begin{table}[t]
\centering
\caption{Task success (\%) on LIBERO ($n{=}500$ per condition). 
ASR pipeline: Whisper ASR + $\pi_{0.5}$ text oracle.
Single LoRA: standard single-adapter LoRA baseline.}
   \vspace*{-7pt}
\label{tab:main}
\setlength{\tabcolsep}{5pt}
\begin{tabular}{@{}lcc@{}}
\toprule
\textbf{Method} & \textbf{Spatial} & \textbf{Object} \\
\midrule
ASR pipeline + $\pi_{0.5}$                 & 95.6 & 95.0 \\
\midrule
Speech-$\pi_{0.5}$ (train voices)       & 89.2 & 86.0 \\
Speech-$\pi_{0.5}$ (held-out accents)   & 85.4 & 84.8 \\
\midrule
\emph{Single LoRA} (train voices)       & 69.0 & 69.0 \\
\emph{Single LoRA} (held-out accents)   & 68.4 & 38.0 \\
\bottomrule
\end{tabular}
\end{table}

\begin{table}[t]
\centering
\caption{Response latency: the latency from speech input to robot action starting (H100, batch=1, 100 runs). 
}
   \vspace*{-7pt}
\label{tab:latency}
\setlength{\tabcolsep}{5pt}
\begin{tabular}{@{}lr@{}}
\toprule
\textbf{Pipeline} & \textbf{First action (ms)} \\
\midrule
ASR pipeline: Whisper-v3 decode     & 216.7 \\
\quad\quad + $\pi_{0.5}$ inference  & 107.7 \\
\quad\quad = total                  & 324.4 \\
\midrule
Speech-$\pi_{0.5}$ (ours)          & 109.1 \\
\midrule
\textbf{Response speedup} & \textbf{3.0$\times$} \\
\bottomrule
\end{tabular}
\end{table}

\section*{Conclusion}

We present Speech-$\pi_{0.5}$, an audio-language-action model that maps speech
directly to robot actions.
Our end-to-end approach eliminates the ASR decoder bottleneck,
achieving $3\times$ lower latency than the traditional ASR decoder pipeline with competitive accuracy. 

\section*{Acknowledgments}

This research is supported in part by grants NSF CMMI 2402438.

{\small
\bibliographystyle{ACM-Reference-Format}
\bibliography{references}
}

\end{document}